\documentclass[11pt,a4paper]{article}
\usepackage[hyperref]{acl2018}
\usepackage{times}
\usepackage{latexsym}
 
\usepackage{url}

\usepackage{graphicx}
\usepackage{multirow}
\usepackage{amsthm}
\usepackage{amsfonts}
\usepackage{color}
\usepackage{MnSymbol}
\usepackage{makecell}
\usepackage{arydshln}
\usepackage{microtype}
\usepackage{amsmath}
\usepackage{algorithm}
\usepackage[noend]{algpseudocode}
\usepackage{algorithmicx}
\usepackage[shortlabels]{enumitem}

\usepackage{caption} 
\usepackage{booktabs}

\newcommand{\ourl}{Low-rank Multimodal Fusion}
\newcommand{\ours}{LMF}

\newcommand{\commentM}[1]{\textcolor{blue}{Modified}}
\definecolor{QRED}{RGB}{169, 53, 43}

\aclfinalcopy 
\def\aclpaperid{1620} 

\DeclareMathOperator*{\myotimes}{\raisebox{-1pt}{\scalebox{1.44}{$\bigotimes$}}}
\DeclareMathOperator*{\myhardamard}{\raisebox{-3.5pt}{\scalebox{1.44}{$\Lambda$}}}
\DeclareMathAlphabet\mathbfcal{OMS}{cmsy}{b}{n}

\title{Efficient Low-rank Multimodal Fusion with Modality-Specific Factors}

\author{}
\author{Zhun Liu$^{*}$, Ying Shen\thanks{$\,\,\,$ equal contributions}$\,\,\,$, Varun Bharadhwaj Lakshminarasimhan, \\ {\bf Paul Pu Liang, Amir Zadeh, Louis-Philippe Morency} \\
School of Computer Science \\
Carnegie Mellon University \\
\texttt { \{zhunl,yshen2,vbl,pliang,abagherz,morency\}@cs.cmu.edu} \\
}
\date{}

\begin{document}
\maketitle

\begin{abstract}

Multimodal research is an emerging field of artificial intelligence, and one of the main research problems in this field is multimodal fusion. The fusion of multimodal data is the process of integrating multiple unimodal representations into one compact multimodal representation. Previous research in this field has exploited the expressiveness of tensors for multimodal representation. However, these methods often suffer from exponential increase in dimensions and in computational complexity introduced by transformation of input into tensor. In this paper, we propose the \ourl\ method, which performs multimodal fusion using low-rank tensors to improve efficiency. We evaluate our model on three different tasks: multimodal sentiment analysis, speaker trait analysis, and emotion recognition. Our model achieves competitive results on all these tasks while drastically reducing computational complexity. Additional experiments also show that our model can perform robustly for a wide range of low-rank settings, and is indeed much more efficient in both training and inference compared to other methods that utilize tensor representations.


\end{abstract}
\section{Introduction}

Multimodal research has shown great progress in a variety of tasks as an emerging research field of artificial intelligence. Tasks such as speech recognition \cite{yuhas1989integration}, emotion recognition, \cite{de1997facial}, \cite{chen1998multimodal}, \cite{wollmer2013youtube}, sentiment analysis, \cite{morency2011towards} as well as speaker trait analysis and media description \cite{park2014computational} have seen a great boost in performance with developments in multimodal research.

However, a core research challenge yet to be solved in this domain is multimodal fusion. The goal of fusion is to combine multiple modalities to leverage the complementarity of heterogeneous data and provide more robust predictions. In this regard, an important challenge has been on scaling up fusion to multiple modalities while maintaining reasonable model complexity. Some of the recent attempts \cite{fukui2016multimodal}, \cite{tensoremnlp17} at multimodal fusion investigate the use of tensors for multimodal representation and show significant improvement in performance. Unfortunately, they are often constrained by the exponential increase of cost in computation and memory introduced by using tensor representations. This heavily restricts the applicability of these models, especially when we have more than two views of modalities in the dataset.

\begin{figure*}[t!]
\centering{
\includegraphics[width=1.0\linewidth]{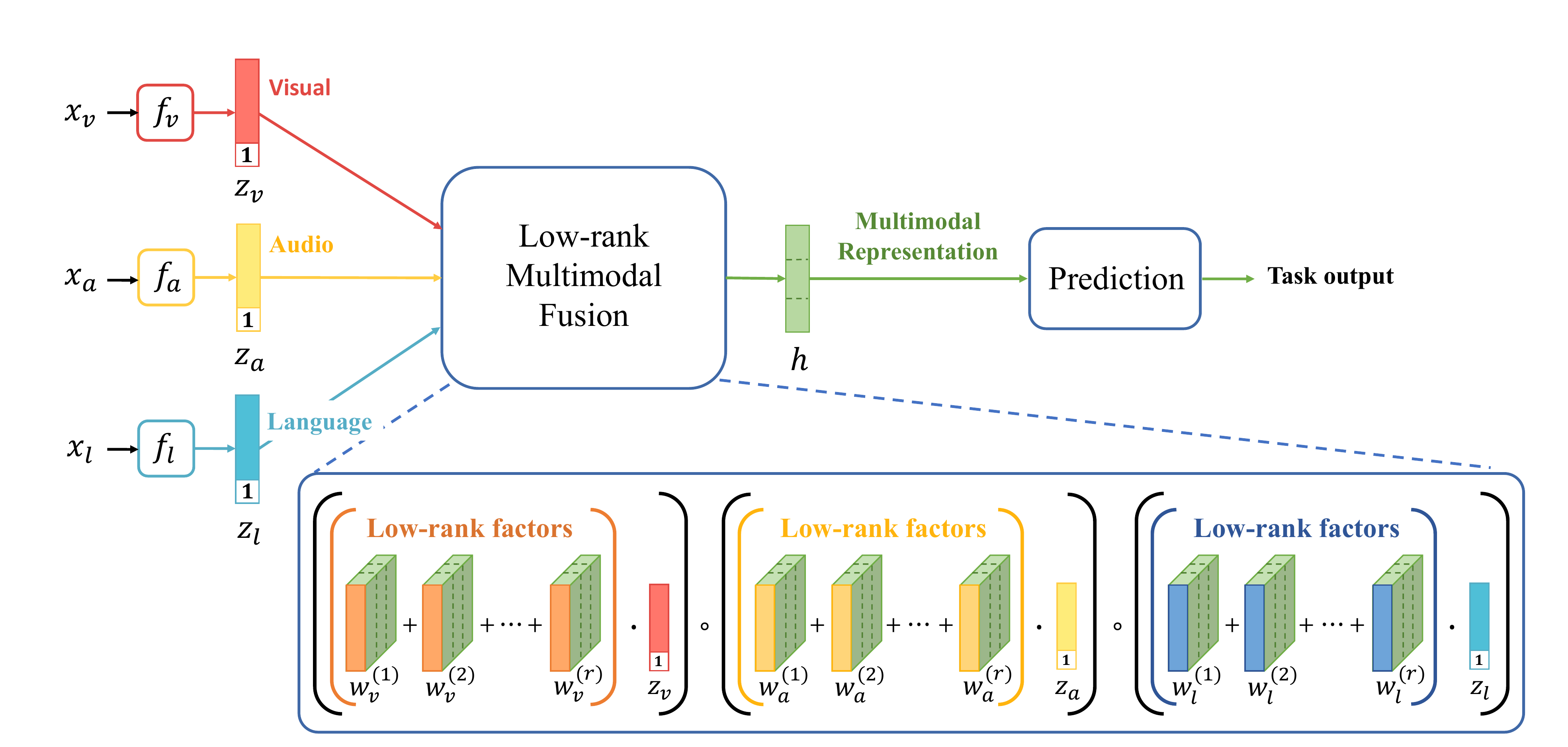}}
\caption{Overview of our \ourl\ model structure: \ours\ first obtains the unimodal representation $z_a, z_v, z_l$ by passing the unimodal inputs $x_a, x_v, x_l$ into three sub-embedding networks $f_v, f_a, f_l$ respectively. \ours\ produces the multimodal output representation by performing low-rank multimodal fusion with modality-specific factors. The multimodal representation can be then used for generating prediction tasks.}
\label{fig:overview}
\end{figure*}

In this paper, we propose the {\ourl}, a method leveraging low-rank weight tensors to make multimodal fusion efficient without compromising on performance. The overall architecture is shown in Figure \ref{fig:overview}. We evaluated our approach with experiments on three multimodal tasks using public datasets and compare its performance with state-of-the-art models. We also study how different low-rank settings impact the performance of our model and show that our model performs robustly within a wide range of rank settings. Finally, we perform an analysis of the impact of our method on the number of parameters and run-time with comparison to other fusion methods. Through theoretical analysis, we show that our model can scale linearly in the number of modalities, and our experiments also show a corresponding speedup in training when compared with other tensor-based models.

The main contributions of our paper are as follows:
\begin{itemize}
\item We propose the {\ourl} method for multimodal fusion that can scale linearly in the number of modalities.
\item We show that our model compares to state-of-the-art models in performance on three multimodal tasks evaluated on public datasets.
\item We show that our model is computationally efficient and has fewer parameters in comparison to previous tensor-based methods.
\end{itemize}
\section{Related Work}

Multimodal fusion enables us to leverage complementary information present in multimodal data, thus discovering the dependency of information on multiple modalities. Previous studies have shown that more effective fusion methods translate to better performance in models, and there's been a wide range of fusion methods.


Early fusion is a technique that uses feature concatenation as the method of fusion of different views. Several works that use this method of fusion \cite{poria2016convolutional} , \cite{wang2016select} use input-level feature concatenation and use the concatenated features as input, sometimes even removing the temporal dependency present in the modalities \cite{morency2011towards}. The drawback of this class of method is that although it achieves fusion at an early stage, intra-modal interactions are potentially suppressed, thus losing out on the context and temporal dependencies within each modality. 

On the other hand, late fusion builds separate models for each modality and then integrates the outputs together using a method such as majority voting or weighted averaging \cite{wortwein2017really}, \cite{nojavanasghari2016deep}. Since separate models are built for each modality, inter-modal interactions are usually not modeled effectively. 

Given these shortcomings, more recent work focuses on intermediate approaches that model both intra- and inter-modal dynamics. \citet{fukui2016multimodal} proposes to use Compact Bilinear Pooling over the outer product of visual and linguistic representations to exploit the interactions between vision and language for visual question answering. Similar to the idea of exploiting interactions, \citet{tensoremnlp17} proposes Tensor Fusion Network, which computes the outer product between unimodal representations from three different modalities to compute a tensor representation. These methods exploit tensor representations to model inter-modality interactions and have shown a great success. However, such methods suffer from exponentially increasing computational complexity, as the outer product over multiple modalities results in extremely high dimensional tensor representations.

For unimodal data, the method of low-rank tensor approximation has been used in a variety of applications to implement more efficient tensor operations. \citet{razenshteyn2016weighted} proposes a modified weighted version of low-rank approximation, and \citet{koch2010dynamical} applies the method towards temporally dependent data to obtain low-rank approximations. As for applications, \citet{lei2014low} proposes a low-rank tensor technique for dependency parsing while \citet{wang2008tensor} uses the method of low-rank approximation applied directly on multidimensional image data (Datum-as-is representation) to enhance computer vision applications. \citet{hu2017attribute} proposes a low-rank tensor-based fusion framework to improve the face recognition performance using the fusion of facial attribute information. However, none of these previous work aims to apply low-rank tensor techniques for multimodal fusion.

Our \ourl\ method provides a much more efficient method to compute tensor-based multimodal representations with much fewer parameters and computational complexity. The efficiency and performance of our approach are evaluated on different downstream tasks, namely sentiment analysis, speaker-trait recognition and emotion recognition.
\section{\ourl}
\label{model}
In this section, we start by formulating the problem of multimodal fusion and introducing fusion methods based on tensor representations. Tensors are powerful in their expressiveness but do not scale well to a large number of modalities. Our proposed model decomposes the weights into low-rank factors, which reduces the number of parameters in the model. This decomposition can be performed efficiently by exploiting the parallel decomposition of low-rank weight tensor and input tensor to compute tensor-based fusion. Our method is able to scale linearly with the number of modalities.

\subsection{Multimodal Fusion using Tensor Representations}
\label{par:stupid_tensor}
In this paper, we formulate multimodal fusion as a multilinear function $f: V_1 \times V_2 \times...\times V_M \rightarrow H$ where $V_1, V_2, ..., V_M$ are the vector spaces of input modalities and $H$ is the output vector space. Given a set of vector representations, $\{z_m\}_{m=1}^M$ which are encoding unimodal information of the $M$ different modalities, the goal of multimodal fusion is to integrate the unimodal representations into one compact multimodal representation for downstream tasks.

Tensor representation is one successful approach for multimodal fusion. It first requires a transformation of the input representations into a high-dimensional tensor and then mapping it back to a lower-dimensional output vector space. Previous works have shown that this method is more effective than simple concatenation or pooling in terms of capturing multimodal interactions \cite{tensoremnlp17}, \cite{fukui2016multimodal}. Tensors are usually created by taking the outer product over the input modalities. In addition, in order to be able to model the interactions between any subset of modalities using one tensor, \citet{tensoremnlp17} proposed a simple extension to append $1$s to the unimodal representations before taking the outer product. The input tensor $\mathcal{Z}$ formed by the unimodal representation is computed by:
\begin{equation}\label{eqn:tensor_prod}
    \mathcal{Z} = \myotimes_{m = 1}^{M} z_m, z_m \in \mathbb{R}^{d_m}
\end{equation}
where $\myotimes_{m = 1}^{M}$ denotes the tensor outer product over a set of vectors indexed by $m$, and $z_m$ is the input representation with appended $1$s.

\begin{figure}[t!]
\centering{
\includegraphics[width=1.0\linewidth]{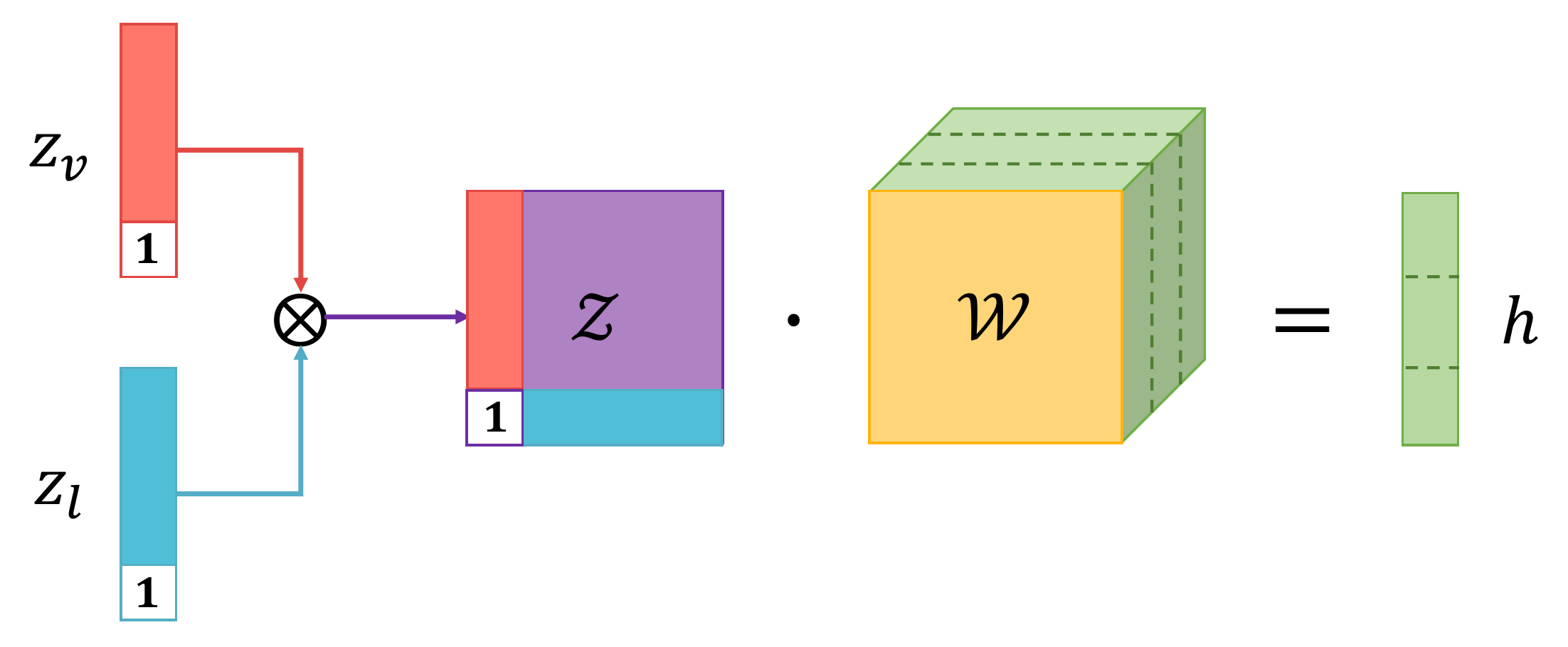}}
\caption{Tensor fusion via tensor outer product}
\label{fig:tfn}
\end{figure}

The input tensor $\mathcal{Z} \in \mathbb{R}^{d_1 \times d_2 \times...d_M}$ is then passed through a linear layer $g(\cdot)$ to to produce a vector representation:
\begin{equation}\label{eqn:linear}
    h = g(\mathcal{Z} ; \mathcal{W}, b) = \mathcal{W} \cdot \mathcal{Z} + b,\ h, b \in \mathbb{R}^{d_y}
\end{equation}
where $\mathcal{W}$ is the weight of this layer and $b$ is the bias. With $\mathcal{Z}$ being an order-$M$ tensor (where $M$ is the number of input modalities), the weight $\mathcal{W}$ will naturally be a tensor of order-$(M+1)$ in $\mathbb{R}^{d_1 \times d_2 \times...\times d_M \times d_h}$. The extra $(M+1)$-th dimension corresponds to the size of the output representation $d_h$. In the tensor dot product $\mathcal{W} \cdot \mathcal{Z}$, the weight tensor $\mathcal{W}$ can be then viewed as $d_h$ order-$M$ tensors. In other words, the weight $\mathcal{W}$ can be partitioned into $\widetilde{\mathcal{W}}_k \in \mathbb{R}^{d_1 \times ... \times d_M},\ k=1,...,d_h$. Each $\widetilde{\mathcal{W}_k}$ contributes to one dimension in the output vector $h$, i.e. $h_k = \widetilde{\mathcal{W}}_k \cdot \mathcal{Z}$.
This interpretation of tensor fusion is illustrated in Figure \ref{fig:tfn} for the bi-modal case.

One of the main drawbacks of tensor fusion is that we have to explicitly create the high-dimensional tensor $\mathcal{Z}$. The dimensionality of $\mathcal{Z}$ will increase exponentially with the number of modalities as $\prod_{m=1}^M d_m$. The number of parameters to learn in the weight tensor $\mathcal{W}$ will also increase exponentially. This not only introduces a lot of computation but also exposes the model to risks of overfitting. 

\subsection{Low-rank Multimodal Fusion with Modality-Specific Factors}
As a solution to the problems of tensor-based fusion, we propose \ourl\ (\ours). \ours\ parameterizes $g(\cdot)$ from Equation \ref{eqn:linear} with a set of modality-specific low-rank factors that can be used to recover a low-rank weight tensor, in contrast to the full tensor $\mathcal{W}$. Moreover, we show that by decomposing the weight into a set of low-rank factors, we can exploit the fact that the tensor $\mathcal{Z}$ actually decomposes into $\{z_m\}_{m=1}^M$, which allows us to directly compute the output $h$ without explicitly tensorizing the unimodal representations. \ours\ reduces the number of parameters as well as the computation complexity involved in tensorization from being exponential in $M$ to linear.

\subsubsection{Low-rank Weight Decomposition}
\label{par:low_rank}
The idea of \ours\ is to decompose the weight tensor $\mathcal{W}$ into $M$ sets of modality-specific factors. However, since $\mathcal{W}$ itself is an order-$(M+1)$ tensor, commonly used methods for decomposition will result in $M+1$ parts. Hence, we still adopt the view introduced in Section \ref{par:stupid_tensor} that $\mathcal{W}$ is formed by $d_h$ order-$M$ tensors $\widetilde{\mathcal{W}}_k \in \mathbb{R}^{d_1 \times ... \times d_M}, k=1,...,d_h$ stacked together. We can then decompose each $\widetilde{\mathcal{W}}_k$ separately.

For an order-$M$ tensor $\widetilde{\mathcal{W}}_k \in \mathbb{R}^{d_1 \times ... \times d_M}$, there always exists an exact decomposition into vectors in the form of:
\begin{equation}\label{eqn:tensor_decomp}
    \widetilde{\mathcal{W}}_k = \sum_{i=1}^{R} \myotimes_{m=1}^M w^{(i)}_{m, k},\ w^{(i)}_{m, k} \in \mathbb{R}^d_m
\end{equation}
The minimal $R$ that makes the decomposition valid is called the \textbf{rank} of the tensor. The vector sets $\{\{w^{(i)}_{m, k}\}_{m=1}^M\}_{i=1}^R$ are called the rank $R$ decomposition factors of the original tensor. 

In \ours, we start with a fixed rank $r$, and parameterize the model with $r$ decomposition factors $\{\{w^{(i)}_{m, k}\}_{m=1}^M\}_{i=1}^r, k=1,...,d_h$ that can be used to reconstruct a low-rank version of these $\widetilde{\mathcal{W}}_k$.

\begin{figure*}[t!]
\centering{
\includegraphics[width=1.0\linewidth]{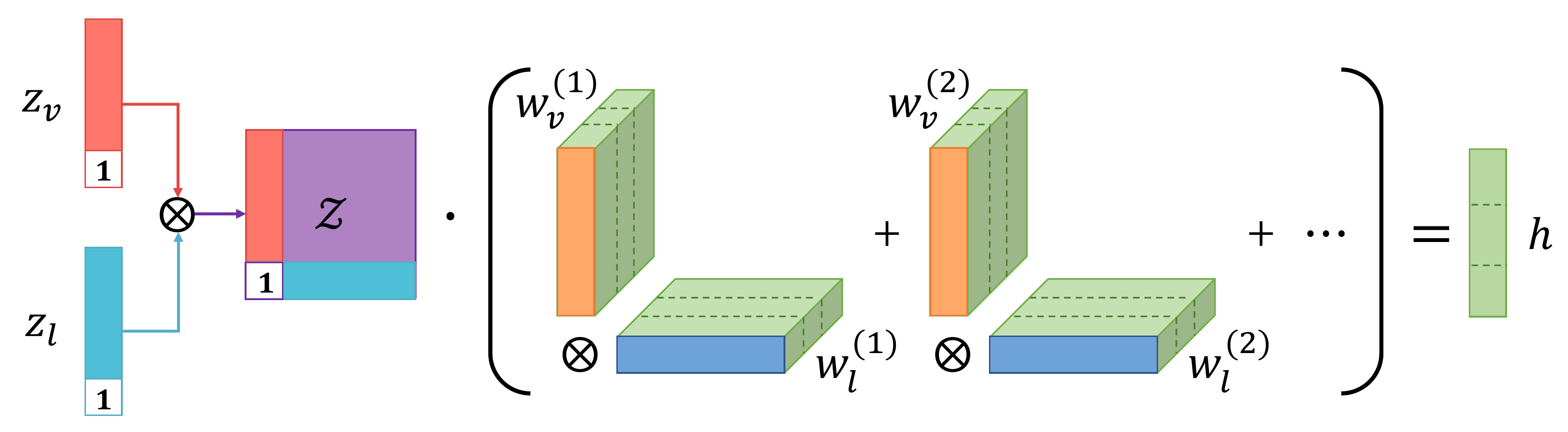}}
\caption{Decomposing weight tensor into low-rank factors (See Section \ref{par:low_rank} for details.)}
\label{fig:lr}
\end{figure*}

We can regroup and concatenate these vectors into $M$ modality-specific low-rank factors. Let $\mathbf{w}_m^{(i)} = [w_{m,1}^{(i)}, w_{m,2}^{(i)},...,w_{m,d_h}^{(i)}]$, then for modality $m$, $\{\mathbf{w}_m^{(i)}\}_{i=1}^r$ is its corresponding low-rank factors. And we can recover a low-rank weight tensor by:
\begin{equation}\label{fig:recon}
\mathcal{W} = \sum_{i=1}^{r} \myotimes_{m=1}^M \mathbf{w}^{(i)}_{m}
\end{equation}
Hence equation \ref{eqn:linear} can be computed by
\begin{align}\label{eqn:naive_lowrank}
h & = \left(\sum_{i=1}^{r} \myotimes_{m=1}^M \mathbf{w}^{(i)}_{m}\right) \cdot \mathcal{Z}
\end{align}
Note that for all $m,\ \mathbf{w}_m^{(i)} \in \mathbb{R}^{d_m \times d_h}$ shares the same size for the second dimension. We define their outer product to be over only the dimensions that are not shared: $\mathbf{w}_m^{(i)} \otimes \mathbf{w}_n^{(i)} \in \mathbb{R}^{d_m \times d_n \times d_h}$.
A bimodal example of this procedure is illustrated in Figure \ref{fig:lr}.


Nevertheless, by introducing the low-rank factors, we now have to compute the reconstruction of $\mathcal{W} = \sum_{i=1}^{r} \myotimes_{m=1}^M \mathbf{w}^{(i)}_{m}$ for the forward computation. Yet this introduces even more computation.

\subsubsection{Efficient Low-rank Fusion Exploiting Parallel Decomposition}
\label{par:clever_tensor}
In this section, we will introduce an efficient procedure for computing $h$, exploiting the fact that tensor $\mathcal{Z}$ naturally decomposes into the original input $\{z_m\}_{m=1}^M$, which is parallel to the modality-specific low-rank factors. In fact, that is the main reason why we want to decompose the weight tensor into $M$ modality-specific factors.

Using the fact that $\mathcal{Z} = \myotimes_{m=1}^M z_m$, we can simplify equation \ref{eqn:naive_lowrank}:
\begin{align}\label{eqn:efficient}
h &= \left( \sum_{i=1}^r \myotimes_{m=1}^M \mathbf{w}_m^{(i)} \right) \cdot \mathcal{Z} \nonumber \\
& = \sum_{i=1}^r \left( \myotimes_{m=1}^M \mathbf{w}_m^{(i)} \cdot \mathcal{Z} \right) \nonumber \\
& = \sum_{i=1}^r \left( \myotimes_{m=1}^M \mathbf{w}_m^{(i)} \cdot \myotimes_{m=1}^M z_m \right) \nonumber \\
& = \myhardamard_{m=1}^M \left[\sum_{i=1}^r \mathbf{w}_{m}^{(i)} \cdot z_m\right]
\end{align}
where $\myhardamard_{m=1}^M$ denotes the element-wise product over a sequence of tensors: $\myhardamard_{t=1}^3 x_t = x_1 \circ x_2 \circ x_3$.


 An illustration of the trimodal case of equation \ref{eqn:efficient} is shown in Figure \ref{fig:overview}. We can also derive equation \ref{eqn:efficient} for a bimodal case to clarify what it does:

\begin{align}\label{eqn:bimodal}
h & = \left( \sum_{i=1}^r \mathbf{w}_a^{(i)} \otimes \mathbf{w}_v^{(i)} \right) \cdot \mathcal{Z} \nonumber \\
& = \left( \sum_{i=1}^r\mathbf{w}_a^{(i)} \cdot z_a \right) \circ \left( \sum_{i=1}^r\mathbf{w}_v^{(i)} \cdot z_v \right)
\end{align}

An important aspect of this simplification is that it exploits the parallel decomposition of both $\mathcal{Z}$ and $\mathcal{W}$, so that we can compute $h$ without actually creating the tensor $\mathcal{Z}$ from the input representations $z_m$. In addition, different modalities are decoupled in the simplified computation of $h$, which allows for easy generalization of our approach to an arbitrary number of modalities. Adding a new modality can be simply done by adding another set of modality-specific factors and extend Equation \ref{eqn:bimodal}. Last but not least, Equation \ref{eqn:efficient} consists of fully differentiable operations, which enables the parameters $\{\mathbf{w}_m^{(i)}\}_{i=1}^r\, m=1,...,M$ to be learned end-to-end via back-propagation.

Using Equation \ref{eqn:efficient}, we can compute $h$ directly from input unimodal representations and their modal-specific decomposition factors, avoiding the weight-lifting of computing the large input tensor $\mathcal{Z}$ and $\mathcal{W}$, as well as the $r$ linear transformation. Instead, the input tensor and subsequent linear projection are computed implicitly together in Equation \ref{eqn:efficient}, and this is far more efficient than the original method described in Section \ref{par:stupid_tensor}. Indeed, \ours\ reduces the computation complexity of tensorization and fusion from $O(d_y \prod_{m=1}^M d_m)$ to $O(d_y \times r \times \sum_{m=1}^M d_m)$.


In practice, we use a slightly different form of Equation \ref{eqn:efficient}, where we concatenate the low-rank factors into $M$ order-3 tensors and swap the order in which we do the element-wise product and summation:
\begin{align}\label{eqn:regression2}
        h =  \sum_{i=1}^r \left[\myhardamard_{m=1}^M \left[ \mathbf{w}_m^{(1)}, \mathbf{w}_m^{(2)}, ..., \mathbf{w}_m^{(r)} \right] \cdot \hat{z}_m\right]_{i, :}
\end{align}
and now the summation is done along the first dimension of the bracketed matrix. $[\cdot]_{i, :}$ indicates the $i$-th slice of a matrix. In this way, we can parameterize the model with $M$ order-3 tensors, instead of parameterizing with sets of vectors.
\section{Experimental Methodology}
\label{experiments}

We compare {\ours} with previous state-of-the-art baselines, and we use the Tensor Fusion Networks (TFN) \cite{tensoremnlp17} as a baseline for tensor-based approaches, which has the most similar structure with us except that it explicitly forms the large multi-dimensional tensor for fusion across different modalities.

We design our experiments to better understand the characteristics of {\ours}. Our goal is to answer the following four research questions:

\noindent\textbf{(1) Impact of Multimodal Low-rank Fusion}: Direct comparison between our proposed {\ours} model and the previous TFN model. 

\noindent\textbf{(2) Comparison with the State-of-the-art}: We evaluate the performance of {\ours} and state-of-the-art baselines on three different tasks and datasets. 

\noindent\textbf{(3) Complexity Analysis}: We study the modal complexity of {\ours} and compare it with the TFN model.

\noindent\textbf{(4) Rank Settings}: We explore performance of {\ours} with different rank settings.

The results of these experiments are presented in Section \ref{results}.

\subsection{Datasets}
\begin{table}[!t]
\fontsize{7}{10}\selectfont
\setlength\tabcolsep{14pt}
\begin{tabular}{l : c : c : c }
\Xhline{0.5\arrayrulewidth}
Dataset   & \multicolumn{1}{c:}{CMU-MOSI}  &  \multicolumn{1}{c:}{IEMOCAP} & \multicolumn{1}{c}{POM} \\
Level       & \multicolumn{1}{c:}{Segment} & \multicolumn{1}{c:}{Segment} & \multicolumn{1}{c}{Video} \\ \Xhline{0.5\arrayrulewidth}
\# Train	& 1284 & 6373 & 600\\  
\# Valid	& 229  & 1775 & 100\\  
\# Test	& 686   &  1807 & 203\\  \Xhline{0.5\arrayrulewidth}
\end{tabular}
\caption{The speaker independent data splits for training, validation, and test sets.}
\label{table:splits}
\end{table}

We perform our experiments on the following multimodal datasets, CMU-MOSI \cite{zadeh2016mosi}, POM \cite{Park:2014:CAP:2663204.2663260}, and IEMOCAP \cite{Busso2008IEMOCAP:Interactiveemotionaldyadic} for sentiment analysis, speaker traits recognition, and emotion recognition task, where the goal is to identify speakers emotions based on the speakers' verbal and nonverbal behaviors.

\noindent\textbf{CMU-MOSI} 
The CMU-MOSI dataset is a collection of 93 opinion videos from YouTube movie reviews. Each video consists of multiple opinion segments and each segment is annotated with the sentiment in the range [-3,3], where -3 indicates highly negative and 3 indicates highly positive. 

\noindent\textbf{POM}
The POM dataset is composed of 903 movie review videos. Each video is annotated with the following speaker traits: confident, passionate, voice pleasant, dominant, credible, vivid, expertise, entertaining, reserved, trusting, relaxed, outgoing, thorough, nervous, persuasive and humorous. 

\noindent\textbf{IEMOCAP} 
The IEMOCAP dataset is a collection of 151 videos of recorded dialogues, with 2 speakers per session for a total of 302 videos across the dataset. Each segment is annotated for the presence of 9 emotions (angry, excited, fear, sad, surprised, frustrated, happy, disappointed and neutral). 

To evaluate model generalization, all datasets are split into training, validation, and test sets such that the splits are speaker independent, i.e., no identical speakers from the training set are present in the test sets. Table \ref{table:splits} illustrates the data splits for all datasets in detail.

\subsection{Features}
Each dataset consists of three modalities, namely language, visual, and acoustic modalities. To reach the same time alignment across modalities, we perform word alignment using P2FA \cite{P2FA} which allows us to align the three modalities at the word granularity. We calculate the visual and acoustic features by taking the average of their feature values over the word time interval \cite{Chen:2017:MSA:3136755.3136801}.

\noindent\textbf{Language} We use pre-trained 300-dimensional Glove word embeddings \cite{pennington2014glove} to encode a sequence of transcribed words into a sequence of word vectors.

\noindent\textbf{Visual} The library Facet\footnote{goo.gl/1rh1JN} is used to extract a set of visual features for each frame (sampled at 30Hz) including 20 facial action units, 68 facial landmarks, head pose, gaze tracking and HOG features \cite{zhu2006fast}.

\noindent\textbf{Acoustic} We use COVAREP acoustic analysis framework \cite{degottex2014covarep} to extract a set of low-level acoustic features, including 12 Mel frequency cepstral coefficients (MFCCs), pitch, voiced/unvoiced segmentation, glottal source, peak slope, and maxima dispersion quotient features.

\subsection{Model Architecture}
In order to compare our fusion method with previous work, we adopt a simple and straightforward model architecture \footnote{The source code of our model is available on Github at https://github.com/Justin1904/Low-rank-Multimodal-Fusion } for extracting unimodal representations. Since we have three modalities for each dataset, we simply designed three unimodal sub-embedding networks, denoted as $f_a, f_v, f_l$, to extract unimodal representations $z_a, z_v, z_l$ from unimodal input features $x_a, x_v, x_l$. For acoustic and visual modality, the sub-embedding network is a simple 2-layer feed-forward neural network, and for language modality, we used an LSTM \cite{Hochreiter:1997:LSM:1246443.1246450} to extract representations. The model architecture is illustrated in Figure \ref{fig:overview}.

\subsection{Baseline Models}
We compare the performance of {\ours} to the following baselines and state-of-the-art models in multimodal sentiment analysis, speaker trait recognition, and emotion recognition.

\noindent\textbf{Support Vector Machines} Support Vector Machines (SVM) \cite{cortes1995support} is a widely used non-neural classifier. This baseline is trained on the concatenated multimodal features for classification or regression task \cite{perez2013utterance}, \cite{park2014computational}, \cite{zadeh2016multimodal}.

\noindent\textbf{Deep Fusion} The Deep Fusion model (DF) \cite{nojavanasghari2016deep} trains one deep neural model for each modality and then combine the output of each modality network with a joint neural network.

\noindent\textbf{Tensor Fusion Network} The Tensor Fusion Network (TFN) \cite{tensoremnlp17} explicitly
models view-specific and cross-view dynamics by creating a multi-dimensional tensor that captures unimodal, bimodal and trimodal interactions across three modalities.

\noindent\textbf{Memory Fusion Network} The Memory Fusion Network (MFN) \cite{zadeh2018memory} accounts for view-specific and cross-view interactions and continuously models them through time with a special attention mechanism and summarized through time with a Multi-view Gated Memory.

\noindent\textbf{Bidirectional Contextual LSTM} The Bidirectional Contextual LSTM (BC-LSTM)  \cite{tensoremnlp17}, \cite{fukui2016multimodal} performs context-dependent fusion of multimodal data.

\noindent\textbf{Multi-View LSTM} The Multi-View LSTM (MV-LSTM) \cite{rajagopalan2016extending} aims to capture both modality-specific and cross-modality interactions from multiple modalities by partitioning the memory cell and the gates corresponding to multiple modalities.

\noindent\textbf{Multi-attention Recurrent Network} The Multi-attention Recurrent Network (MARN) \cite{zadeh2018multi} explicitly models interactions between modalities through time using a neural component called the Multi-attention Block (MAB) and storing them in the hybrid memory called the Long-short Term Hybrid Memory (LSTHM).

\subsection{Evaluation Metrics}
Multiple evaluation tasks are performed during our evaluation: multi-class classification and regression. The multi-class classification task is applied to all three multimodal datasets, and the regression task is applied to the CMU-MOSI and the POM dataset. For binary classification and multiclass classification, we report F1 score and accuracy Acc$-k$ where k denotes the number of classes. Specifically, Acc$-2$ stands for the binary classification. For regression, we report Mean Absolute Error (MAE) and Pearson correlation (Corr). Higher values denote better performance for all metrics except for MAE.
\section{Results and Discussion}
\label{results}

In this section, we present and discuss the results from the experiments designed to study the research questions introduced in section \ref{experiments}.

\subsection{Impact of \ourl}
In this experiment, we compare our model directly with the TFN model since it has the most similar structure to our model, except that TFN explicitly forms the multimodal tensor fusion.
The comparison reported in the last two rows of Table \ref{table:performance} demonstrates that our model significantly outperforms TFN across all datasets and metrics. This competitive performance of {\ours} compared to TFN emphasizes the advantage of \ourl.

\subsection{Comparison with the State-of-the-art}

\begin{table*}[th!]
\fontsize{7.5}{10}\selectfont
\setlength\tabcolsep{2.1pt}
\begin{tabular}{l : *{5}{p{1cm}<{\centering}}: *{3}{p{1cm}<{\centering}}: *{4}{p{1.2cm}<{\centering}}}
\Xhline{0.5\arrayrulewidth}
Dataset & \multicolumn{5}{c:}{\textbf{CMU-MOSI}} & \multicolumn{3}{c:}{\textbf{POM}} & \multicolumn{4}{c}{\textbf{IEMOCAP}}  \\
Metric    & MAE & Corr & Acc-2 & F1 & Acc-7 & MAE & Corr & Acc &  F1-Happy & F1-Sad & F1-Angry & F1-Neutral \\ 
\Xhline{0.5\arrayrulewidth}
SVM & 1.864 & 0.057 & 50.2 & 50.1 & 17.5 & 0.887 & 0.104 & 33.9 & 81.5 &78.8& 82.4& 64.9 \\
DF    & 1.143   &  0.518  & 72.3 &   72.1   &  26.8 & 0.869 & 0.144 & 34.1 & 81.0 & 81.2 & 65.4 & 44.0 \\
BC-LSTM    & 1.079 &  0.581     & 73.9 &   73.9   &  28.7& 0.840 & 0.278 & 34.8	& 81.7 & 81.7 & 84.2 & 64.1  \\ 
MV-LSTM		& 1.019 &  0.601	& 73.9 &   74.0   &  33.2	& 0.891 & 0.270 & 34.6 & 81.3 & 74.0 & 84.3 & 66.7 \\
MARN  & 0.968  & 0.625 & 77.1	& 77.0 & 34.7 & -    &  -  & 39.4 & 83.6 & 81.2 & 84.2 & 65.9 \\ 
MFN  	&  {0.965}    	& {0.632} & {\textbf{77.4}}	 & {\textbf{77.3}}   & {\textbf{34.1}} & 0.805   & 0.349 & 41.7 & 84.0 & 82.1 & 83.7 & 69.2 \\
\Xhline{1\arrayrulewidth}
TFN  	& 0.970	& 0.633   & 73.9 &   73.4   &  32.1 & 0.886 & 0.093  & 31.6 & 83.6 &82.8 & 84.2 & 65.4 \\ 
{\ours} & \textbf{0.912} &	\textbf{0.668} & 76.4& 75.7	& 32.8 & \textbf{0.796} & \textbf{0.396} & \textbf{42.8} & \textbf{85.8} & \textbf{85.9 }& \textbf{89.0} & \textbf{71.7 }\\
\Xhline{0.5\arrayrulewidth}
\end{tabular}
\caption{
Results for sentiment analysis on CMU-MOSI, emotion recognition on IEMOCAP and personality trait recognition on POM. Best results are highlighted in bold.
}
\label{table:performance}
\end{table*}

We compare our model with the baselines and state-of-the-art models for sentiment analysis, speaker traits recognition and emotion recognition. Results are shown in Table \ref{table:performance}. {\ours} is able to achieve competitive and consistent results across all datasets.

On the multimodal sentiment regression task, {\ours} outperforms the previous state-of-the-art model on MAE and Corr. Note the multiclass accuracy is calculated by mapping the range of continuous sentiment values into a set of intervals that are used as discrete classes. 

On the multimodal speaker traits Recognition task, we report the average evaluation score over 16 speaker traits and shows that our model achieves the state-of-the-art performance over all three evaluation metrics on the POM dataset.

On the multimodal emotion recognition task,  our model achieves better results compared to the state-of-the-art models across all emotions on the F1 score. F1-emotion in the evaluation metrics indicates the F1 score for a certain emotion class.

\subsection{Complexity Analysis}
Theoretically, the model complexity of our fusion method is $O(d_y \times r \times \sum_{m=1}^M d_m)$ compared to $O(d_y \prod_{m=1}^M d_m)$ of TFN from Section \ref{par:stupid_tensor}. In practice, we calculate the total number of parameters used in each model, where we choose $M=3$, $d_1 = 32$, $d_2 = 32$, $d_3 = 64$, $r = 4$, $d_y = 1$. Under this hyper-parameter setting, our model contains about 1.1e6 parameters while TFN contains about 12.5e6 parameters, which is nearly 11 times more. Note that, the number of parameters above counts not only the parameters in the multimodal fusion stage but also the parameters in the subnetworks.

\begin{table}[!tbp]
\fontsize{7}{10}\selectfont
\setlength\tabcolsep{14pt}
\begin{tabular}{l:c:c}
\Xhline{0.5\arrayrulewidth}
{Model} & {Training Speed (IPS)} & {Testing Speed (IPS)} 
\\
\Xhline{0.5\arrayrulewidth}
TFN &  340.74 &   1177.17     \\
\ours & 1134.82 &   2249.90      \\ 
\Xhline{0.5\arrayrulewidth}
\end{tabular}
\caption{Comparison of the training and testing speeds between TFN and {\ours}. The second and the third columns indicate the number of data point inferences per second (IPS) during training and testing time respectively. Both models are implemented in the same framework with equivalent running environment.
}
\label{table:running_time}
\end{table}
Furthermore, we evaluate the computational complexity of {\ours} by measuring the training and testing speeds between {\ours} and TFN. Table \ref{table:running_time} illustrates the impact of {\ourl} on the training and testing speeds compared with TFN model. Here we set rank to be 4 since it can generally achieve fairly competent performance. 

Based on these results, performing a low-rank multimodal fusion with modality-specific low-rank factors significantly reduces the amount of time needed for training and testing the model. On an NVIDIA Quadro K4200 GPU, {\ours} trains with an average frequency of 1134.82 IPS (data point inferences per second) while the TFN model trains at an average of 340.74 IPS. 


\subsection{Rank Settings}
\begin{figure}[b!]
\centering{
\includegraphics[width=1.05\linewidth]{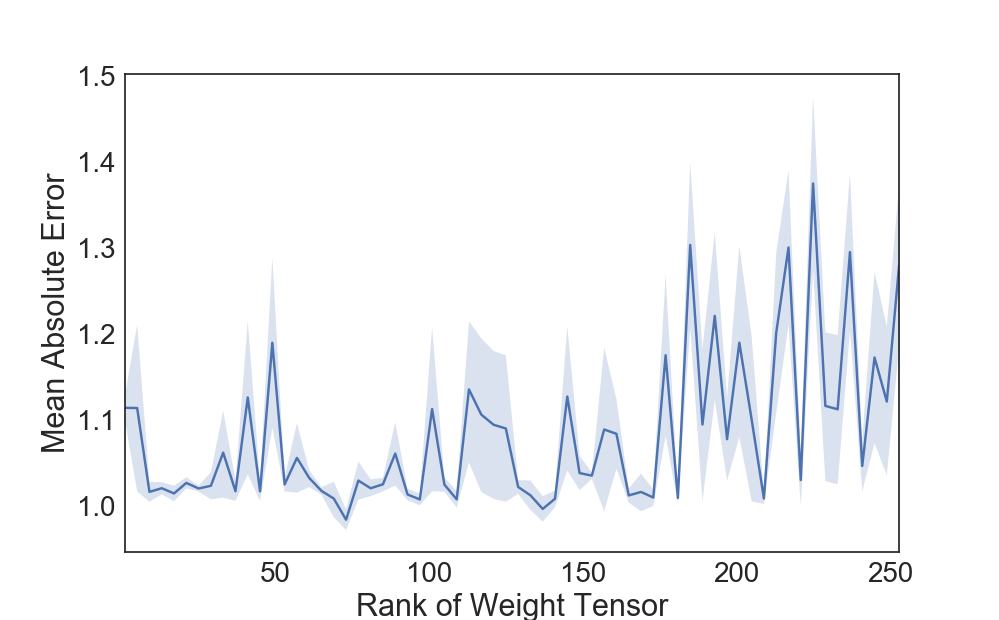}}
\caption{The Impact of different rank settings on Model Performance: As the rank increases, the results become unstable and low rank is enough in terms of the mean absolute error.
}
\label{fig:rank}
\end{figure}

To evaluate the impact of different rank settings for our {\ours} model, we measure the change in performance on the CMU-MOSI dataset while varying the number of rank. The results are presented in Figure \ref{fig:rank}. We observed that as the rank increases, the training results become more and more unstable and that using a very low rank is enough to achieve fairly competent performance.
\section{Conclusion}
In this paper, we introduce a {\ourl} method that performs multimodal fusion with modality-specific low-rank factors. {\ours} scales linearly in the number of modalities. {\ours} achieves competitive results across different multimodal tasks. Furthermore, {\ours} demonstrates a significant decrease in computational complexity from exponential to linear time. In practice, {\ours} effectively improves the training and testing efficiency compared to TFN which performs multimodal fusion with tensor representations.

Future work on similar topics could explore the applications of using low-rank tensors for attention models over tensor representations, as they can be even more memory and computationally intensive.

\section*{Acknowledgements}
This material is based upon work partially supported by the National Science Foundation (Award \# 1833355) and Oculus VR. Any opinions, findings, and conclusions or recommendations expressed in this material are those of the author(s) and do not necessarily reflect the views of National Science Foundation or Oculus VR, and no official endorsement should be inferred.

\bibliography{acl2018_bib}
\bibliographystyle{acl_natbib}

\end{document}